\documentclass{article}

\usepackage{arxiv}

\usepackage{graphicx}
\usepackage{wrapfig}

\usepackage[table,pdflatex]{xcolor}

\usepackage{pgf}
\newcommand\inputpgf[2]{{
		\let\pgfimageWithoutPath\pgfimage
		\renewcommand{\pgfimage}[2][]{\pgfimageWithoutPath[##1]{#1/##2}}
		\input{#1/#2}
}}

\usepackage{booktabs}
\usepackage{etoolbox}
\usepackage{microtype}

\usepackage{siunitx}
\graphicspath{{images/}}

\usepackage[unicode=true,breaklinks=true,colorlinks=false,linkcolor=black,pdfborder={0 0 0},pdfdisplaydoctitle=true,pdftitle={Using anomaly detection to support classification of fast running packaging processes},pdfauthor={Tilman Klaeger}]{hyperref}

\hypersetup{final}

\title{	Using anomaly detection to support classification of fast running (packaging) processes}

\author{
	\textbf{Tilman Klaeger, Andre Schult, Lukas Oehm\thanks{© 2019 IEEE.  Personal use of this material is permitted.  Permission from IEEE must be obtained for all other uses, in any current or future media, including reprinting/republishing this material for advertising or promotional purposes, creating new collective works, for resale or redistribution to servers or lists, or reuse of any copyrighted component of this work in other works. \textbf{Final version published under DOI \href{http://doi.org/10.1109/INDIN41052.2019.8972081}{10.1109/INDIN41052.2019.8972081}}}}\\
	Fraunhofer Institute for Process Engineering and Packaging (IVV)\\
		Division Machinery and Processes\\
		Dresden, Germany \\
		{tilman.klaeger@ivv-dresden.fraunhofer.de}
}

\begin{document}

\maketitle

\begin{abstract}
	
In this paper we propose a new method to assist in labeling data arriving from fast running processes using anomaly detection. A result is the possibility to manually classify data arriving at a high rates to train machine learning models. To circumvent the problem of not having a real ground truth we propose specific metrics for model selection and validation of the results. 

The use case is taken from the food packaging industry, where processes are affected by regular but short breakdowns causing interruptions in the production process. Fast production rates make it hard for machine operators to identify the source and thus the cause of the breakdown. Self learning assistance systems can help them finding the root cause of the problem and assist the machine operator in applying lasting solutions. These learning systems need to be trained to identify reoccurring problems using data analytics. Training is not easy as the process is too fast to be manually monitored to add specific classifications on the single data points.	
\end{abstract}


\section{Introduction}

Self learning systems like operator assistance need valid training data.
For slow running processes it is possible to manually label failed production steps or poorly manufactured products.
For fast running discrete processes this is much harder as it is difficult to manually distinguish the single products. 

Packaging processes are one of those fast running discrete processes. Packing sweets runs at up to 2\,000 pieces per minute, filling tubes at a rate of 600 per minute \cite{Majschak2018AnwendungfuerMaschinensysteme}. The processing of materials with volatile properties like plastic foil and cardboard or treating of biogene products, e.g. foods, causes frequent interruptions in the production process \cite{Schult2015SteigerungEffizienzVerarbeitungs, Tsarouhas2012Reliabilityavailabilitymaintainability}. Research at the Fraunhofer IVV using data gained in 7\,000 hours of manual machine observation has shown an average production time frame between errors being less than 15 minutes (Mean time between failure, MTBF). But it has also to be mentioned that in 66\% of the interruptions the fix needs less than two minutes (Mean time to repair, MTTR) \cite{Dziuba-Kaiser2017ErmittlungWiederholcharakteristikim}. The overall equipment efficiency (OEE) is not only affected by interruptions but also by producing with less quality than required. This will cause rejects or, in the worst case, food safety issues due to incorrect packaging. To refer to both breakdowns and non-quality production the term "defect" is used in this paper.

A major cause of the frequent interruptions is the missing ability of operators to understand the problem and fix the problem's root cause. Building assistance systems can help operators to find lasting solutions for the problem \cite{Rahm2018KoMMDiaDialogueDrivenAssistance, Muller2018Processindustriesdiscrete}. To build more sophisticated views of the current failure situation an analysis of all available PLC signals is used to build a self learning assistance system \cite{Klaeger2017LernfahigeBedienerassistenzfur}.

For first experiments lasting malfunctions have been provocated over a period of a couple minutes to create training data collected from the PLC for machine learning algorithms \cite{Klaeger2017LernfahigeBedienerassistenzfur}. This is a simplified approach not matching up with real production scenarios where often only a single product or cardboard sheet may have a problem and either results in non-quality products or causes a machine stop. So there is a need to identify situations for every single product being produced. And more challenging, a possibility to classify the correct data point after manually reporting some kind of malfunction needs to be implemented.

In this paper we propose a way using anomaly detection to identify and assist in manual labeling the data points arriving at a high rate. We prove the performance by using a new kind of metric helping to find the best combination of feature selection, anomaly detection and a final classification.

\section{Related work}

\subsection{Data classification for machine state detection on packaging machines}

Many researches are working in the field of predictive maintenance to tackle down-times before an important machine component fails. The focus is on preventing longer machine stops. A major research in this area is the analysis of different kind of rotating machine parts like motors, bearings and pumps \cite{Lei2016IntelligentFaultDiagnosis, Lauro2014Monitoringprocessingsignal}. Most analysis is done in the frequency domain, even though some have tried working in the time domain \cite{Rapur2017ExperimentalTimeDomainVibrationBased}.

Due to intermittent motions in many parts of the packaging machine analysis in the frequency domain is not always possible and thus a inspection of the time domain has to be done \cite{Lotze2008SteuerungsintegrierteVorgangsdiagnoseVerarbeitungsmaschinen, Behrendt2009AntriebsbasierteZustandsuberwachungam}. 

For continuous movements in packaging machines there have been successful analysis in the frequency domain. Brecher et al. are using the vibration measurements to detect deterioration of drive belts \cite{Brecher2017Optimizedstateestimation}. The experimental setup allowed for longer running experiments using drive belts at different wear levels.

Carino et al. perform analysis in the frequency domain working with motor currents for a camshaft of a packaging machine to detect different production states \cite{Carino2015Noveltydetectionmethodology}. Deactivating sections of the machine to create different states of operation eases the creation of a classification. The same setup is also used by Zurita et al. \cite{Zurita2015Diagnosismethodbased} to identify states using artificial neural networks.

Without the application of machine learning Ostyn et al. were able to detect the quality of pouch sealings using statistical analyses of multivariate accelerometer data measured at the sealing jaw \cite{Ostyn2007StatisticalMonitoringSealing}. Running at about 30 bags per minute a manual inspection of the seal integrity was possible.

Opposed to most research our focus is to tackle short but frequent defects in discrete processes having summarized a major share on unplanned machine stops. The method has to be able to detect defects in single products or machine cycles.

\subsection{Anomaly detection, model selection and evaluation of anomaly detectors}

Instead of combining anomaly detection and classification one could look completely separated at those methods. Classification of PLC data is possible as shown in \cite{Klaeger2017LernfahigeBedienerassistenzfur}.

For anomaly detection there is a variety of established algorithms. Having classified nominal data at hand a semi-supervised approach is possible, sometimes refereed to as novelty detection. Available methods can basically be divided in different categories \cite{Goldstein2016ComparativeEvaluationUnsupervised}. Global neighbor based algorithms based on \textit{k}-Nearest Neighbor (\textit{k}-NN) are somewhat related to other density-based algorithms like the Local Outlier Factor \cite{Breunig2000LOFIdentifyingDensitybased}. There are algorithms available based on subspace methods like Principal Component Analysis (PCA) or based on statistics like Histogram Based Outlier Detection (HBOS) \cite{Goldstein2012HistogrambasedOutlierScore} or Gaussian Mixture Models. Further algorithms include Isolation Forests based on decision tree methods and others based on the Minimum Covariance Distance (MCD) \cite{Hardin2004Outlierdetectionmultiple}.
All of those can be used with feature bagging methods, that are well established for classification tasks but are also suitable for outlier analysis \cite{Aggarwal2017OutlierAnalysis}. Except for Gaussian Mixture all those algorithms are available in the open library "PyOD" \cite{Zhao2019PyODPythonToolbox}. 

Performance evaluation is the crucial point when selecting appropriate algorithms for anomaly detection. Common metrics for bench-marking anomaly detection are the receiver operating characteristic (ROC), the area under the curve (AUC) or simple metrics like precision and recall. Those metrics are often used, especially for comparison of different models \cite{Chandola2008ComparativeEvaluationAnomaly, Goldstein2016ComparativeEvaluationUnsupervised}. But these metrics always need a ground truth to work \cite{Aggarwal2017OutlierAnalysis, Xu2019EnsembleLensEnsemblebasedVisual}. If no ground truth is available characterizing the performance of anomaly detection is difficult and often internal metrics are used \cite{Aggarwal2017OutlierAnalysis}.

\section{Proposed Method}

\subsection{Machine used for experiments and data collection}

\begin{figure*}

	\centering
	\begin{footnotesize}
	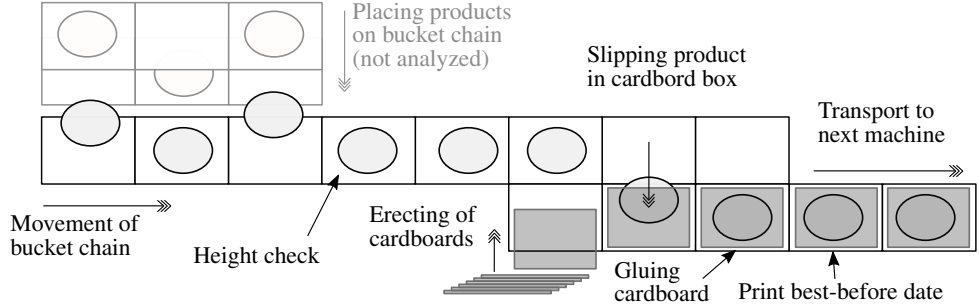
	\end{footnotesize}

	\caption{Illustration of the analyzed packaging process.}
	\label{fig:SketchProcess}
\end{figure*}

The data was collected at a cartoner in a live production environment. The machine is built to pack up to 120 products per minute. Working at variable speeds the machine can adjust to the filling level at the preceding production line. The machine is an off-the-shelf cartoner retrofitted at the food producers' site. The camshaft and other parts of the machine are now equipped with independent servo drives and new PLC. For a visualization of the process see figure \ref{fig:SketchProcess}.

Incoming products are placed in one of three collators and then dropped on the next free spot on the bucket chain. For further research only the processes running alongside the bucket chain have been analyzed. Products in the bucket chain are subject to different handling and checks:

\begin{enumerate}
	\item Checking for product height, unintended stacking of products and correct placement in bucket.
	\item Detracting and erecting of cardboard boxes with two open sites.
	\item Slipping product into cardboard box.
	\item Gluing the open sites of the box.
	\item Printing best-before date and checking the print image.
\end{enumerate}

Data from the built in sensor system was collected on the PLC's field bus using the PLC's cycle time of \SI{1}{ms} in live production. A total of 40 sensors was available for analysis including rotary encoders but excluding those of the safety circuit. 

To train the machine learning models the cartoner was observed at three shifts lasting seven hours each. In this time periods of time were logged without any error. Defects, especially those not causing a machine stop, were logged but have to be treated with some uncertainty in observation time. 

The most common defect observed in production was products not positioned correctly in the bucket chain. Most of those were detected by the machine and only rarely caused machine stops by ripped cardboards when trying to package the product. In some cases the detracting of the cardboard failed. This has to occur multiple times as the machine is programmed to retry on missed detraction. Another reasons for down times were stops at the machine following in the packaging line. A list of observed states is shown in table \ref{Tab:ObservedDefects}.

\begin{table}[h]
	\caption{Observed states in the cartoner.}
	\label{Tab:ObservedDefects}
	\centering
	\small
	\begin{tabular}{lr}
		\toprule 
		\textbf{State / Defect} & \textbf{Num. of observations} \\ 
		\midrule 
		Known Nominal data without reported defect & 2.850\\
		\midrule
		Product not placed correctly in bucket chain & 37 \\ 
		 
		Interruption caused by next machine in line& 7 \\ 
		
		Cardboard Detraction error & 6 \\ 
		 
		Failed to slip product in box & 2 \\ 
		 
		Error depositing product in bucket chain & 1 \\ 
		\midrule 
		\textbf{Sum of observed defects} & \textbf{53} \\
		\bottomrule
	\end{tabular} 
\end{table}

For evaluation all defects including those detected by the machine were used. As some defects do have only one or two occurrences in the data set all defects were classified as "defect" for evaluation. Splitting between different failures prevents the use of cross validation techniques.

After having installed the equipment for data collection only very few breakdowns could be observed while watching the machine in production. This does not match prior observations at various other packaging machines \cite{Schult2015SteigerungEffizienzVerarbeitungs, Dziuba-Kaiser2017ErmittlungWiederholcharakteristikim}. Despite this fact the data is of major interest as data from packaging machines in real production is still rare and the general format of data will not change with a more error prone machine.

In addition to data collected while watching the machine personally, data was collected remotely and thus has no classifications. Using machine stops as ground truth for defects is not feasible since stops may be caused by defects as well as by manual stops due to missing packaging goods or other planed down times (e.\,g. cleaning, maintenance).

\subsection{Feature extraction}

There are two basic options to produce equal length data sets: Looking at the machine for a defined amount of time, usually one cycle respectively one turn of the camshaft. Defects caused by anomalous products may then be visible more than once at every sensor having contact with the product. But these impacts may be very minor. To track errors on the product a digital twin was built collecting the sensor data virtually on the product. As result a data set for every product passing through the machine is available.

To overcome different length caused by varying machine speeds a scaling was applied to the collected data sets. To be able to analyze data sets in case of machine stops a timeout is implemented for early finishing of digital twins. Missing data in those data sets is filled using median value imputation.

To extract features from these data sets common figures of time series are calculated \cite{Esling2012Timeseriesdatamining}.
Using process knowledge about the type of sensors the features are selected resulting at an average of five features per sensor and a total of 200. The resulting high dimension can trivially be reduced to 120 by setting a low variance threshold. For a further reduction of the dimension transformation-methods like Principal Component Analysis (PCA) and Factor Analysis (FA) are applied.

\subsection{Building pipelines with anomaly detection and classification}

\begin{figure*}
	
	\centering
	\begin{footnotesize}
		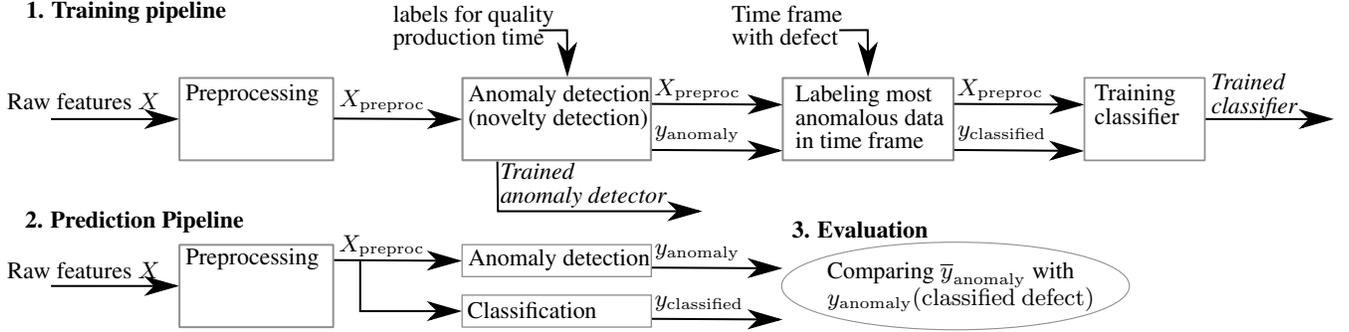
	\end{footnotesize}
	
	\caption{Construction of the pipeline for with anomaly detection and classification for training and prediction.}
	\label{fig:Pipeline}
\end{figure*}

The most important part in classifying fast running processes is to add the user reported classification to the correct data point. For the classification of a single data point we propose a method consisting of anomaly detection and classification: To classify a single data point the anomaly score in a reported time range is used and the data point with the highest score within this time range is labeled with the reported defect. Assuming a good anomaly detection the classified data set can be used as training data for the classifier. The approach and the resulting pipeline is visualized in figure~\ref{fig:Pipeline}.

Looking at the available data not only unsupervised methods for anomaly detection can be considered but also methods for novelty detection using a semi-supervised approach with known nominal training data. For the current research a variety of detectors including HBOS, Isolation Forest, \textit{k}-NN, LOF, MCD and approaches based on PCA and Gaussian Mixtures were used.

As feature bagging methods have shown good results for classification on similar data sets random subspace sampling was added \cite{Klaeger2017LernfahigeBedienerassistenzfur}. The anomaly detectors were provided with 50\% and 70\% of the features using an ensemble size of 80 detectors. For comparison a detector without random subspace sampling was also present for each anomaly detection method.

Combined with different sorts of scaling the data and feature reduction methods and a random subspace sampling for anomaly detectors a total of about 10\,000 variants were built and (semi)-automatically examined in a typical grid search. 

\subsection{Proposed and used metrics}

As a simple evaluation of anomaly detectors is not trivial, especially if there is no ground truth available \cite{Aggarwal2017OutlierAnalysis}. In consequence often internal criteria need to be used. Having a complete pipeline the proposed feature can be called semi-internal: We propose the consensus of anomaly score and classification as metric to validate the performance of a combination. The metric $m$ therefore is based on calculating the multiple of average anomaly score for detected defects $\overline{y}_{\mathrm{defect}}$ compared to the average of all anomaly scores $\overline{y}$. In the total average anomly-score detected defects are included. This works particularly well for data sets with a small fraction of errors, where a couple defects hardly affect the total average. In other cases a comparison between defects and known nominal data is more suitable.

\[	m = \frac{\overline{y}_{\mathrm{defect}}}{\overline{y}} \]

An important remark is that one assumption has to be made at this point: The data point marked using the anomaly detection is identified correctly. Going one step further the assumption should prove wrong if the anomaly detection returns – worst case – random values. A classifier trained with random values will also return random values and thus perform poor in a cross validation test.

Having observed nominal values available and best guesses on data for defects it is further possible to use well-known and proven metrics for data science: Precision, Recall and F1-score. It is possible to calculate the precision for nominal production. Using the assumption from the previous paragraph, that anomaly detection is working correctly and identified the correct data points, it is possible to do a cross validation on the classifier. With these metrics a plausibility check of the first proposed metric is made. 

Other metrics analyzed include the variance of the anomaly detection. As the data set contains a skewed distribution with only very few anomalies, a low variance in the anomaly score is expected. So this may be a metric to determine the quality of anomaly detection. For better comparison between different methods all anomaly scores are scaled to be in range 0 to 100.

Another expectation is an even distribution of defects over time. Thus performing a $\chi^2$-test over the detected defects is yet another metric for determining performance of the model.

\section{Results}

Out of the 10\,000 variations analyzed three were extraordinary. They all used a PCA for feature reduction with the number of components automatically determined with the method by Minka \cite{Minka2001AutomaticChoiceDimensionality}. For anomaly detection MCD with random subspace sampling of 70\% of the features was used. The final classification was done using Random Forests without weighting the classes.

In different metrics other combinations scored slightly higher. The best precision for defects was as high as 100\% and a recall up to 53\% could be achieved using another combination. The maximum needs to be interpreted carefully as it is known that a recall for defects of 100\% is easy to reach by classifying all data points as defect. 

With the best combination examined the average anomaly score for detected defects is 155 times as high as the average anomaly score off all sampled data. This results in a precision of 87\% and a recall of 49\%, both being very close to the maximum achieved results. Poor performance was only measured for the criterion of an even spread of detected defects over time. All results are summarized in table \ref{Tab:ResultMetrics}. 

\begin{table*}[b]
	\centering
	\scriptsize
	\caption{Realized performance with the best combination of anomaly score and classifier compared to overall performance.}
	\label{Tab:ResultMetrics}
	\begin{tabular}{lr|rr|rrr|rrr}
\toprule
	&	&	\multicolumn{2}{c|}{\textbf{Anomalies}} & \multicolumn{3}{c|}{\textbf{Defects}} &  \multicolumn{3}{c}{\textbf{Unweighted Average}}	\\		
 &\textbf{ Rel. Score} &  \textbf{Variance} &  \textbf{$\chi^2$ stats} &  \textbf{F1-score} &  \textbf{Precision} &  \textbf{Recall} & \textbf{F1-score} &  \textbf{Precision} &  \textbf{Recall} \\

\midrule
Best performing Combination            &           155.4 &    0.884 &                    611.4 &         0.627 &          0.867 &       0.491 &            0.811 &             0.929 &          0.745 \\
Relative to best of each  &            1.00 &    0.999 &                      0.020 &       0.906 &          0.867 &       0.928 &            0.960 &             0.932 &          0.974 \\

\midrule
Mean                    &           10.42 &   177.28 &                   2391.4 &         0.367 &          0.865 &       0.254 &            0.680 &             0.926 &          0.626 \\
Standard Derivation                    &           16.02 &   320.58 &                   5905.4 &         0.192 &          0.170 &       0.152 &            0.097 &             0.085 &          0.076 \\
Minimum                     &            0.00 &     0.62 &                      0.0 &         0.000 &          0.000 &       0.000 &            0.495 &             0.490 &          0.499 \\
75\% Quantile                 &           10.54 &   190.92 &                   1007.5 &         0.538 &          1.000 &       0.396 &            0.766 &             0.991 &          0.697 \\
Maximum                     &          155.36 &  1617.87 &                  30494.4 &         0.691 &          1.000 &       0.528 &            0.843 &             0.995 &          0.764 \\
\midrule
Correlation to Rel. Score & 1.00 & -0.218	&  -0.035 & - 0.219 &  -0.011 &  -0.201 & -0.218 & -0.014 & -0.201 \\
\bottomrule
\end{tabular}

\end{table*}

\begin{figure}[h]
		\centering
		\input{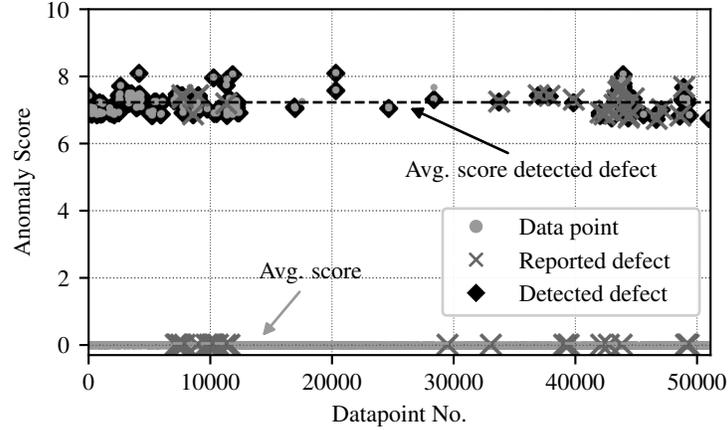}
	\caption{Visualization of calculated anomaly scores with markers for manually reported defects and automatically detected ones.}
	\label{Plt:GesamterProzess}
\end{figure}

A visualization of the detection is displayed in figure \ref{Plt:GesamterProzess}. Only very few data points with high anomaly score are not detected as defect, but some manually reported defects (marked with~$\times$) are not anomaly according to the detector. A recall of 50\% is visible.

First results seam reasonable for the detected metric but do not ensure a transferability to other combinations examined. A common measure is the Pearson correlation factor between the different metrics. Here only negative correlations can be observed which may be suitable for anomaly variance since an overall low anomaly score should result in high model performance. These results need further discussion.

The new metrics are not adoptable to prior experiments (s. \cite{Klaeger2017LernfahigeBedienerassistenzfur}) as defects were acquired continuous. Thus selecting the most anomalous data set close to reported will (nearly) always succeed as neighboring data points contain to the same provocated defect. Until new data sets for different machines are generated there is now further validation possible.

\section{Discussion}

For all metrics the key concern is that our ground truth can not be considered a real ground truth. By quantifying the difference between classifier-detected defects and the average anomaly score for all data points in the set a good common understanding of the situation by anomaly and classification models can be assumed.

A precision of close to 87\% may not always be enough for automatic controls and adaptive systems but it is more than convenient for operator assistance systems. To many false alarms will reduce the benefit and the acceptance of such systems which should not be the case with accomplished performance. A recall at close to 50\% is still motivation for improvement but on the other side does not do too much harm to the systems. For a fully automated system this may not be enough but is beyond the goal as the intention is to provide operator assistance as good as possible and not automatically interfere with the machine.

Having a negative correlation between well-known metrics and the proposed one may reduce the validity of the new approach. Looking at the overall performance of the best performing models the new approach may also be seen as a different metric independent of well-known ones. Using a combination of those metrics it is possible to get a similar result as with the new metric. An automatic model selection is thus easier since only one metric has to be evaluated.

By use of cross validation the metrics for Precision, Recall and F1-score should not be prone to an over fitted model. But having a poor performance on the even distribution of the detected defects is still a slight evidence for over fitting. This issue has to be carefully watched once more data is available.

As of now it is hard to tell how well the new approach adopts to different data sets. The assumptions made for the selection of the metrics do apply for many discrete processes where only few anomal data points are observed.

\section{Conclusion}

We proposed a way to classify data originating from fast running discrete processes using a combination of anomaly detection and classification.  With a new metric model selection was reduced to evaluate one value. 

The search of over 10\,000 model combinations was successful in a way, that it is now possible to detect anomalies and use those for classification of processes where the specific data point can not be manually selected for classification. 

For self learning assistance systems the initial training is thus possible without provocation of lasting errors. The solution thus adds to the economical factor of bringing up such systems at a new machine. But it also helps in working closer on practical data as any kind of provocation has to be seen as simulation specific machine states which may or may not be the same as in real production.

\section*{Acknowledgment}

This research was made possible by funding from the European Regional Development Fund (ERDF) for Saxony, Germany in the project "Lernf{\"a}higes Bediener-Assistenzsystem für Verarbeitungsmaschinen".\\
\includegraphics[width=4.6cm]{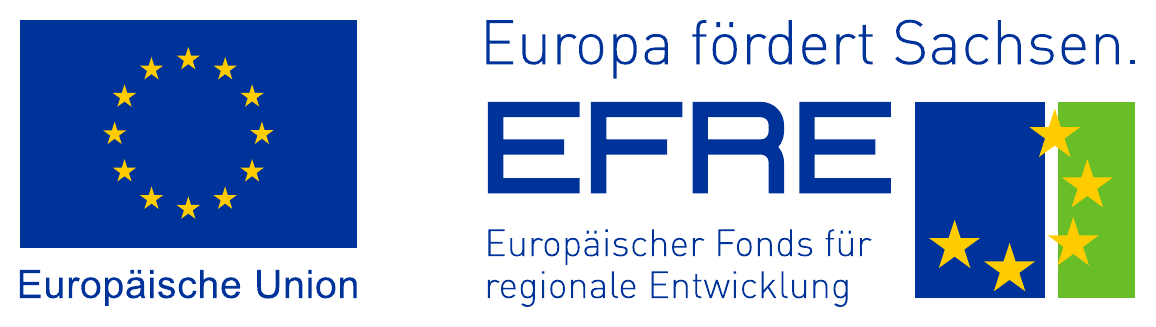}

\newpage

\bibliographystyle{unsrt}
\bibliography{z_citations.bib}

\vfill

\begingroup
\setlength{\intextsep}{0pt}
\setlength{\columnsep}{3pt}
\parindent0pt

\begin{minipage}[l][1.4in]{\textwidth}
\begin{wrapfigure}{l}{1.1in}
\includegraphics[width=1in,height=1.25in,clip,keepaspectratio]{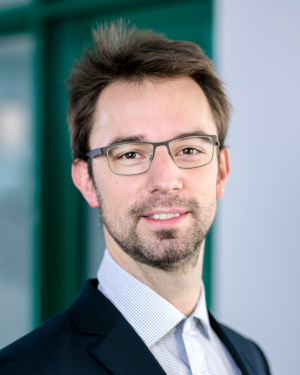}
\end{wrapfigure}\par
\textbf{Dipl.-Ing. Tilman Klaeger} studied mechatronics at the Technische Universit{\"a}t Dresden and is working at the Fraunhofer IVV since 2016. His major topic in research is machine learning on industrial data collected from packaging machines and processes.\par
\end{minipage}

\begin{minipage}[t][1.4in]{\textwidth}
\begin{wrapfigure}{l}{1.1in}
\includegraphics[width=1in,height=1.25in,clip,keepaspectratio]{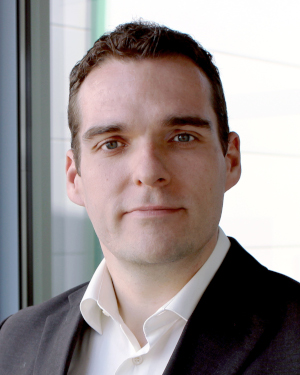}
\end{wrapfigure}
\textbf{Dipl.-Ing. Andre Schult}	studied mechanical engineering with focus on processing and packaging machines at the Technische Universit{\"a}t Dresden. He has been doing research on machine and overall line performance and is now leading the development of user assistance systems. He is currently also founding a company to market assistance systems.
\end{minipage}

\begin{minipage}[t][1.5in]{\textwidth}
\begin{wrapfigure}{l}{1.1in}
\includegraphics[width=1in,height=1.25in,clip,keepaspectratio]{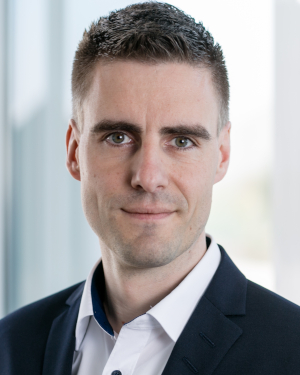}
\end{wrapfigure}
\textbf{Dr.-Ing. Lukas Oehm} graduated in mechanical engineering in 2010 and received his PhD in 2017 from Technische Universit{\"a}t Dresden entitled  “Joining of Polymeric Packaging Materials with High Intensity Focused Ultrasound”. He has been working as research assistant at Fraunhofer Institute for Process Engineering and Packaging IVV since May 2017. Since October 2018, he is group leader for Digitization and Assistance Systems. His research interests are in the field of product safety issues and process efficiency in food production as well as assistance systems for machine operators.
\end{minipage}

\endgroup
\end{document}